\documentclass[a4paper,fleqn]{cas-sc}

\usepackage[authoryear,longnamesfirst]{natbib}
\usepackage{pifont}
\usepackage{lineno}
\usepackage{caption}
\usepackage{tabularx} 
\usepackage{amsmath,amssymb}
\usepackage{algorithm,algpseudocode}
\usepackage{siunitx}             
\usepackage{float}               
\usepackage{pgfplots}
\pgfplotsset{compat=1.18}
\usepackage{empheq}

\def\tsc#1{\csdef{#1}{\textsc{\lowercase{#1}}\xspace}}
\tsc{WGM}
\tsc{QE}
\tsc{EP}
\tsc{PMS}
\tsc{BEC}
\tsc{DE}

\begin{document}
\let\WriteBookmarks\relax
\def\floatpagepagefraction{1}
\def\textpagefraction{.001}

\shorttitle{VLM-Based End-to-End Cooperative Autonomous Driving}

\shortauthors{Junwei You et~al.}

\title [mode = title]{V2X-VLM: End-to-End V2X Cooperative Autonomous Driving Through Large
Vision-Language Models}                      



%
\author[1]{Junwei You}[orcid=0009-0002-6447-8276]






\affiliation[1]{organization={Department of Civil and Environmental Engineering, University of Wisconsin–Madison},
    city={Madison},
    state={WI},
    postcode={53706}, 
    country={USA}}

\affiliation[2]{organization={College of Computing and Data Science, Nanyang Technological University},
    city={Singapore},
    postcode={639798}, 
    country={Singapore}}

\affiliation[3]{organization={College of Transportation, Tongji University},
    city={Shanghai},
    postcode={201804}, 
    country={China}}

\affiliation[4]{organization={Zachry Department of Civil and Environmental Engineering, Texas A\&M University},
    city={College Station},
    state={TX},
    postcode={77840}, 
    country={USA}}

\affiliation[5]{organization={School of Civil and Environmental Engineering, Cornell University},
    city={Ithaca},
    state={NY},
    postcode={14853},
    country={USA}}

\author[2]{Zhuoyu Jiang}

\author[1]{Zilin Huang}

\author[3,1]{Haotian Shi}
\ead{shihaotian95@tongji.edu.cn}
\cormark[1]

\author[1]{Rui Gan}

\author[4]{Keshu Wu}

\author[5]{Xi Cheng}

\author[1]{Xiaopeng Li}

\author[1]{Bin Ran}


\cortext[cor1]{Corresponding author}



\begin{abstract}
Vehicle-to-everything (V2X) cooperation has emerged as a promising paradigm to overcome the perception limitations of classical autonomous driving by leveraging information from both ego-vehicle and infrastructure sensors. However, effectively fusing heterogeneous visual and semantic information while ensuring robust trajectory planning remains a significant challenge. This paper introduces V2X-VLM, a novel end-to-end (E2E) cooperative autonomous driving framework based on vision-language models (VLMs). V2X-VLM integrates multiperspective camera views from vehicles and infrastructure with text-based scene descriptions to enable a more comprehensive understanding of driving environments. Specifically, we propose a contrastive learning-based mechanism to reinforce the alignment of heterogeneous visual and textual characteristics, which enhances the semantic understanding of complex driving scenarios, and employ a knowledge distillation strategy to stabilize training. Experiments on a large real-world dataset demonstrate that V2X-VLM achieves state-of-the-art trajectory planning accuracy, significantly reducing L2 error and collision rate compared to existing cooperative autonomous driving baselines. Ablation studies validate the contributions of each component. Moreover, the evaluation of robustness and efficiency highlights the practicality of V2X-VLM for real-world deployment to enhance overall autonomous driving safety and decision-making.
\end{abstract}


\maketitle

\section{Introduction}

\begin{figure*}[htbp]
    \centering
    \includegraphics[width=\linewidth]{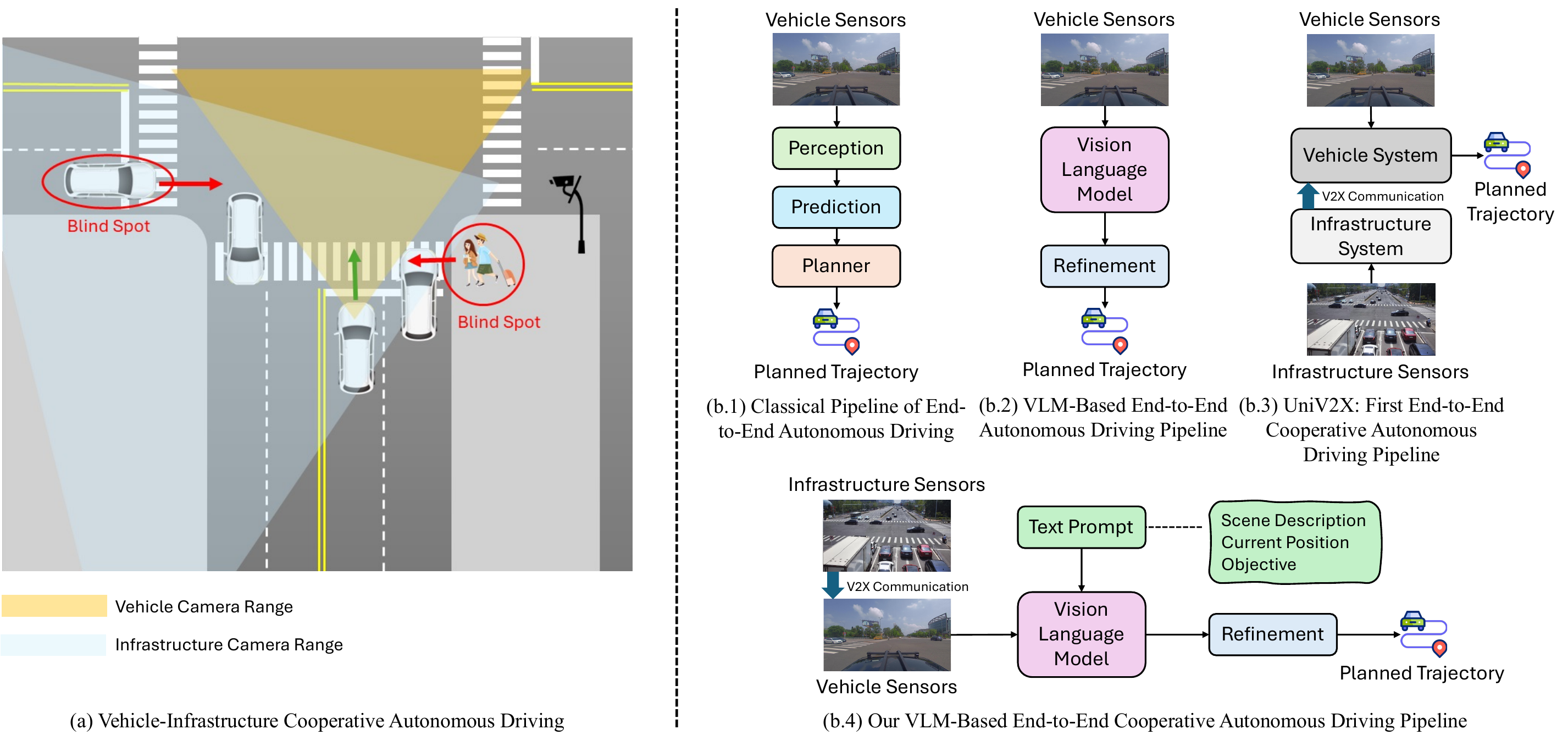}
    \caption{Overview of end-to-end autonomous driving pipelines. (a) A cooperative driving scenario where infrastructure sensors supplement the ego vehicle’s limited field of view; (b.1) the classical end-to-end pipeline that relies solely on on-board sensor data; (b.2) a VLM-based end-to-end system that integrates multimodal reasoning within a single vehicle; (b.3) UniV2X—the pioneering end-to-end cooperative autonomous driving pipeline that fuses vehicle and infrastructure data; and (b.4) our proposed V2X-VLM framework, which leverages large VLM to unify multimodel data for robust end-to-end trajectory planning.}
    \label{fig:fig1}
\end{figure*}

End-to-end (E2E) autonomous driving has emerged as a compelling paradigm by directly mapping raw sensor inputs to vehicle control commands, offering a simplified alternative to labor-intensive modular pipelines (\cite{hu2023planning, jiang2023vad, li2024ego, zheng2024genad}). While these classical E2E methods reduce hand-engineered complexity, they often struggle to interpret complex traffic scenarios without higher-level semantic reasoning. 

Emerging developments in foundation models, especially large language models (LLMs) and vision-language models (VLMs), introduce richer multimodal understanding, which enables E2E pipelines to better interpret visual scenes and textual cues (\cite{xu2024drivegpt4, sima2024drivelm, shao2024lmdrive, tian2024drivevlm, fu2024drive, ma2024dolphins, huang2024vlm, hwang2024emma}). However, since both classical and VLM-enhanced E2E systems rely solely on a single vehicle's sensor data, they remain limited in challenging conditions where supplemental context is needed, such as occlusions and blind spots.

Cooperative autonomous driving extends beyond the single-vehicle view by leveraging Vehicle-to-Everything (V2X) communication to integrate data from both vehicles and infrastructure. As illustrated in Figure~\ref{fig:fig1}(a), infrastructure sensors contribute crucial contextual information that complements the ego vehicle’s field of view. In this case, the early cooperative methods focused mainly on vehicle-to-vehicle (V2V) data fusion (\cite{wang2020v2vnet, cui2022coopernaut, xu2022v2x, hu2023collaboration, xu2024v2x}) to enhance perception under multi-agent settings. Nevertheless, they typically addressed only partial tasks such as detection or occupancy mapping, which fell short of offering fully integrated planning and control, limiting their ability to provide E2E solutions for real-world autonomous driving scenarios. 

To address this problem, more recent studies culminate in UniV2X (\cite{yu2024end}), the first E2E cooperative autonomous driving pipeline that fuses data from both vehicles and infrastructure to produce an integrated framework for comprehensive perception and planning. Figure~\ref{fig:fig1}(b.1-b.3) illustrates the evolution of existing E2E autonomous driving approaches from the classical pipeline to UniV2X-represented cooperative pipeline. However, UniV2X still relies on traditional deep learning architectures that struggle to unify heterogeneous sensor data for complete semantic understanding, especially given the increased complexity introduced by infrastructure-side inputs. These data streams not only capture a broader field of view, but also reflect more intricate road geometries, additional traffic agents, and a wealth of environmental context not observed by onboard vehicle sensors. Thus, there is a pressing need for more advanced approaches capable of bridging these disparate perspectives and extracting cohesive high-level representations to drive a more effective cooperative E2E decision-making.



Inspired by the success of VLMs in single-vehicle E2E setups - where advanced scene understanding and reasoning are achieved through the multimodal fusion of visual data and textual cues (\cite{xu2024drivegpt4, sima2024drivelm, shao2024lmdrive, tian2024drivevlm, fu2024drive, ma2024dolphins, huang2024vlm, hwang2024emma, jiao2024lavida, long2024vlm, liu2025codrivevlm, zhang2024interndrive})—we posit that integrating VLMs in a cooperative autonomous driving framework could enhance joint perception, situation awareness, and planning accuracy. Motivated by this, we propose V2X-VLM, a novel VLM-based E2E cooperative autonomous driving framework, as shown in Figure~\ref{fig:fig1}(b.4). Unlike single-modality E2E architectures (\cite{hu2023planning, jiang2023vad, xu2024v2x}), V2X-VLM unifies multiperspective sensor data from vehicles and infrastructure, augmenting them with text prompts for advanced spatial semantic reasoning. 


Specifically, to further reinforce the cross-modal alignment of contextual and visual information in traffic conditions, we introduce a contrastive learning-based feature alignment mechanism that achieves more discriminative situation awareness for effective trajectory planning. Furthermore, to improve the efficiency of large VLM training, we incorporate a teacher–student distillation strategy during fine-tuning, which produces a smoother learning process. The main contributions are as follows:
\begin{itemize}
    \item We introduce a novel VLM-based end-to-end cooperative autonomous driving framework that unifies sensor data from both vehicles and infrastructure with textual scene descriptions, thus enhancing trajectory planning through advanced multimodal understanding.
    \item We propose a contrastive learning-based alignment mechanism that explicitly synchronizes visual inputs with their corresponding textual cues, resulting in a more discriminative understanding of complex driving scenarios.
    \item We integrate a knowledge distillation strategy during fine-tuning to stabilize the learning process and efficiently transfer rich multimodal representations.
    \item We evaluated the proposed framework on the DAIR-V2X dataset (\cite{yu2022dair}), demonstrating significant improvements over the state-of-the-art methods. The robustness and efficiency evaluation validates the practicality of V2X-VLM for real-world deployment.
\end{itemize}

\section{Related Work}

\subsection{End-to-End Autonomous Driving}
E2E autonomous driving directly maps raw sensor data to vehicle control commands without relying on a fully disassembled perception–prediction–planning pipeline. Common approaches typically used convolutional or transformer-based architectures to infer vehicle movements from onboard camera views (\cite{hu2023planning, jiang2023vad, hu2022st,shao2023reasonnet,ye2023fusionad,chen2024vadv2,sun2024sparsedrive,li2024hydra,guo2024end,yuan2024drama}). These classical frameworks minimize human-engineered components, but often struggle with occlusions and complex multi-agent interactions. More recent lines of work adopt generative models (\cite{zheng2024genad, liao2024diffusiondrive}) to capture uncertainties in future states, or leverage occupancy-based (\cite{mahjourian2022occupancy, li2024end}), Gaussian-based (\cite{zheng2024gaussianad}), or world model-based~(\cite{li2024enhancing, gao2024enhance, wang2024drivedreamer, wang2024driving}) representations to enhance geometric and dynamic understanding. Increasingly, LLMs and VLMs (\cite{xu2024drivegpt4, sima2024drivelm, shao2024lmdrive, tian2024drivevlm, fu2024drive, ma2024dolphins, huang2024vlm, hwang2024emma, jiao2024lavida, long2024vlm, liu2025codrivevlm, zhang2024interndrive}) have also been integrated to inject richer semantic cues and enable higher-level reasoning. In parallel, the field distinguishes between open-loop approaches, which generate future trajectories or control commands without feedback, and closed-loop approaches that continuously update actions in real time. Most of the aforementioned paradigms remain open-loop, although some recent methods have attempted closed-loop integration, typically under simulated or controlled conditions (\cite{huang2024vlm, wang2024driving, jia2025bench2drive,shao2024lmdrive,huang2024human}).

Compared to these paradigms, our proposed V2X-VLM capitalizes on multiperspective data from vehicles and infrastructure to enhance the onboard sensor view, enhancing spatio-temporal coverage especially in occluded or visually ambiguous situations. Unlike conventional single-modality or single-vehicle pipelines, V2X-VLM unifies heterogeneous visual streams with textual prompts in a large VLM backbone, providing robust collaborative semantic reasoning for complex road scenarios, and thus resulting in promising motion planning outcomes.

\subsection{Cooperative Autonomous Driving}
Cooperative autonomous driving leverages V2V and vehicle-to-infrastructure (V2I) communication to integrate distributed sensing and decision making. Although early work primarily focused on the fusion of data between multiple vehicles to improve perception quality (\cite{xu2022cobevt,yu2023v2x,lu2023robust,chen2023transiff,hu2022where2comm,cui2022coopernaut}), recent efforts emphasize infrastructure-based sensing that broadens the field of view for enhanced situational awareness. This shift reflects the limitation of V2V fusion, which is constrained by shared occlusions and low viewpoints. Infrastructure-side sensing provides a top-down perspective of complex road networks and captures broader traffic dynamics involving diverse participants (\cite{yi2024v2iviewer, mo2024enhanced, khan2025vehicle}). For example, at occluded intersections or curved on-ramps, infrastructure cameras can observe vehicles and pedestrians outside the line of sight of onboard sensors. Similarly, in dense urban traffic with multi-lane merges, infrastructure input enables a more complete understanding of agent interactions beyond what nearby vehicles can perceive alone. UniV2X (\cite{yu2024end}) represents a critical step, which integrates vehicle and infrastructure data into an E2E pipeline for comprehensive perception and planning. 


Yet, UniV2X still relies on traditional deep learning models that often struggle with the semantic richness and heterogeneity of multi-perspective data streams. In contrast, V2X-VLM addresses these limitations of UniV2X by unifying heterogeneous vehicle and infrastructure data within a large VLM backbone, augmented by textual prompts that inject high-level contextual cues. This design enables stronger semantic grounding, improved robustness to occlusions, and better generalization across diverse traffic scenarios, resulting in more accurate and reliable E2E planning.




\section{Problem Formulation}
The objective of the proposed V2X-VLM framework is to plan an optimal trajectory for the ego vehicle by leveraging heterogeneous sensor data from both vehicles and infrastructure, along with textual prompts that inject high-level semantic context.

Concretely, let $I_v \in \mathbb{R}^{H_v \times W_v \times 3}$ represent the camera input of the ego vehicle of height $H_v$ and width $W_v$, $I_i \in \mathbb{R}^{H_i \times W_i \times 3}$ denote the image data from the infrastructure cameras of height $H_i$ and width $W_i$, and $E$ signify the textual prompt that contains contextual signals; our objective is to predict a discrete sequence of 2D positions for the ego vehicle over a time horizon $T$, producing a trajectory $\tau$:
\begin{linenomath}
\begin{equation}
\tau = \{(x_t, y_t) \;\mid\; t = 1, 2, \ldots, T \},
\end{equation}
\end{linenomath}
where $(x_t, y_t)$ denotes the planned location in the 2D ground plane at time $t$. Our end-to-end model $F(\cdot)$ learns to generate $\tau$ directly from the inputs $(I_v, I_i, E)$, trained by minimizing the discrepancy between the predicted trajectory $\tau$ and the ground truth $\tau^*$. Formally, we solve the following:

\begin{linenomath}
\begin{equation}
\min \mathcal{L}(\tau, \tau^*) \;=\; \min \mathcal{L}\bigl(F(I_v, I_i, E), \tau^*\bigr),
\end{equation}
\end{linenomath}
where $\mathcal{L}(\cdot)$ is a suitable loss function. By fusing multiperspective visual data with textual prompts, V2X-VLM obtains a robust semantic understanding of the driving environment, and thus enables the direct prediction of the ego vehicle’s movement in complex and real-world scenarios.

\section{Methodology}

\subsection{V2X-VLM Framework}
\label{sec:method} 
\begin{figure*}[t]
    \centering
    \includegraphics[width=\textwidth]{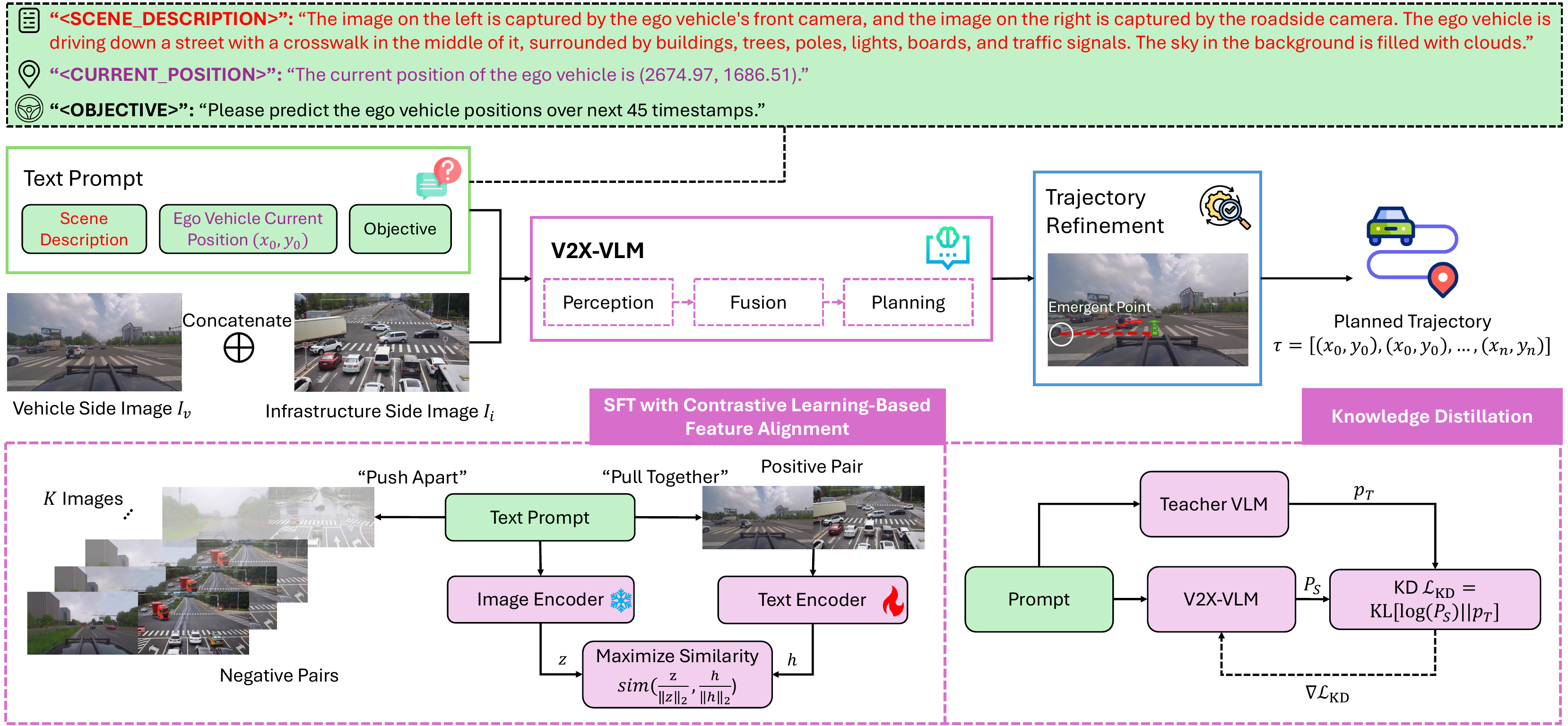} 
    \caption{\textbf{Overview of the Proposed V2X-VLM Framework.} Camera images from the vehicle and infrastructure sides merged with semantic text prompt are fed in a VLM backbone for multiperspective and multimodel data fusion. Through comprehensive scene understanding and reasoning, V2X-VLM delivers accurate and reliable E2E trajectory planning. Contrastive learning-based feature alignment is applied during fine tuning to ensure the effective fusion of visual and semantic features for enhanced scene understanding. Knowledge distillation stabilizes the learning process to fulfill the complex E2E autonomous driving task.}
    \label{fig:framework2}
\end{figure*}

The overall framework of V2X-VLM is demonstrated in Figure~\ref{fig:framework2}. 
As addressed previously, V2X-VLM generates ego-vehicle trajectories by fusing heterogeneous visual inputs from vehicles and infrastructure, together with textual prompts that provide high-level semantic context.

Specifically, the vehicle camera image $I_v$ captures critical real-time information about the vehicle's local surroundings, while infrastructure-side image $I_i$, collected from cameras placed at strategic points like intersections, provides a wider view of broader traffic patterns and pedestrian activities that might not be visible from the vehicle's perspective. In addition to visual data, the framework incorporates a text prompt $E$, which includes semantic textual information relevant to the driving context. It encompasses three key elements: scene description resulted from the ability of VLM to understand and interpret the complex driving environment and crafted by human; the current position of the ego vehicle serving as the planning basis; as well as the explicit planning task description. These inputs are fed into a large VLM backbone containing a visual encoder for $I_v$ and $I_i$ and a text encoder for $E$. The outputs of these encoders are merged into a shared latent space. This process allows for a more synthetic analysis of the environment, where visual cues and textual information are correlated to provide a holistic understanding of the situation. The primary output of V2X-VLM is a planned trajectory $\tau$ for the ego vehicle. 

To further enhance the correct correlation between image inputs and text descriptions, a contrastive learning-based feature alignment technique is employed during fine tuning. Furthermore, a knowledge distillation strategy is leveraged to ensure efficient and stabilized and training process with knowledge transferring. Both methods are detailed in the following sections. The trajectory refinement procedure is also applied to avoid the planning result being skewed by misleading or atypical data points.

\subsection{Multimodal Feature Alignment}
Contrastive learning-based technique is conducted to align visual features from \((I_v, I_i)\) with text features from \(E\). This alignment ensures that the model accurately correlates each visual scene with its corresponding semantic description, thereby strengthening robust scene understanding (\cite{zeng2023clip2, liu2023contrastive}).

\paragraph{Feature Extraction.}
Given a vehicle image \(I_v\) and an infrastructure image \(I_i\), we first concatenate them along the width for an image tensor \([I_v, I_i] \in \mathbb{R}^{H \times (W_v + W_i) \times 3}\). This composite image is then processed by the image encoder within VLM, and its output is aggregated via a pooling function to produce a fixed length visual embedding:
\[
z \;=\; \mathrm{pooling}\Bigl(\mathrm{image\_encoder}\bigl([I_v,I_i]\bigr)\Bigr) \;\in\;\mathbb{R}^{d_z}.
\]
Simultaneously, the text encoder processes the textual input \(E\), and a pooling operation yields the textual embedding:
\[
h \;=\; \mathrm{pooling}\Bigl(\mathrm{text\_encoder}(E)\Bigr) \;\in\;\mathbb{R}^{d_h}.
\]
For simplicity, we align the dimensions so that \(d_z = d_h = d'\), ensuring both embeddings are compatible for subsequent multimodal alignment.

\paragraph{Contrastive Alignment.}
We apply \(\ell_2\) normalization to both \(z\) and \(h\):
\begin{equation}
    \hat{z} \;=\; \frac{z}{\|z\|_2}, 
    \quad
    \hat{h} \;=\; \frac{h}{\|h\|_2}.
\end{equation}
Given a training batch of size \(K\), we compute pairwise similarities \(S_{ij}\) for \(i,j \in \{1,\dots,K\}\) as:
\begin{equation}
    S_{ij}
    \;=\;
    \frac{\hat{z}_i^\top \,\hat{h}_j}{\kappa},
\end{equation}
where \(\kappa\) is a temperature hyperparameter controlling the sharpness of the similarity distribution. Positive and correct image-text pairs \((i = i)\) are encouraged to have larger similarity scores \(S_{ii}\), while negative and incorrect pairs \((i \neq j)\) are penalized, as shown in Figure~\ref{fig:framework2}. By doing so, each visual embedding \(\hat{z}_i\) is brought close to its matching text embedding \(\hat{h}_i\), and pushed away from all unrelated text embeddings. This approach improves the understanding of the heterogeneous scene of the V2X-VLM framework by ensuring that the combined multiperspective image aligns correctly with its corresponding prompt. Matching the image with the correct prompt adds an additional layer of validation, which further refines the model's understanding of traffic scenes beyond the processing capabilities of the VLM alone.

\subsection{Knowledge Distillation}
Training such a large VLM with diverse cooperative data from multiple cameras and textual prompts for outstanding performance can be challenging. To efficiently transfer multimodal knowledge while stabilizing the training dynamics, we employ a teacher-student distillation strategy (\cite{hinton2015distilling,wang2022efficientvlm,zhang2024vlm}) with temperature scaling. As shown in Figure~\ref{fig:framework2}, we maintain a frozen pretrained teacher model \(F_T\) and a trainable student model \(F_S\) initialized with pretrained weights. Both models process identical input batches \((I_v, I_i, E)\), producing trajectory logits \(\tau_T = F_T(I_v, I_i, E)\) and \(\tau_S = F_S(I_v, I_i, E)\), respectively.

\paragraph{Softened Distribution Matching.} 
We calculate the KL divergence between the student's predictions and the teacher's temperature-scaled distribution. First, we soften both logits with a temperature parameter \(\mathcal{T}\):

\begin{linenomath}
\begin{equation}
    \tau'_T = \frac{\tau_T}{\mathcal{T}}, \quad \tau'_S = \frac{\tau_S}{\mathcal{T}}.
\end{equation}
\end{linenomath}

The teacher's target probabilities are then obtained via softmax normalization:

\begin{linenomath}
\begin{equation}
    p_T = \mathrm{softmax}(\tau'_T).
\end{equation}
\end{linenomath}

The student's log-probabilities are obtained as following:

\begin{linenomath}
\begin{equation}
    \log p_S = \mathrm{log\_softmax}(\tau'_S).
\end{equation}
\end{linenomath}

\paragraph{Distillation Loss Formulation.} 
The final KL divergence loss encourages distributional alignment between student and teacher:

\begin{linenomath}
\begin{equation}
    \mathcal{L}_{\text{KD}} = \mathcal{T}^2 \cdot \mathrm{KL}\left(\log p_S \,\|\, p_T\right).
\end{equation}
\end{linenomath}

The \(\mathcal{T}^2\) multiplier compensates for gradient scaling induced by temperature, ensuring stable optimization. This softened target distribution provides richer supervision than hard labels, particularly during early training, when the student's random initialization leads to unstable gradients.

\subsection{Training Objective}
\label{sec:training} 
The complete training objective of V2X-VLM combines three key components:

\paragraph{Trajectory Prediction Loss.} 
The primary trajectory prediction loss in the context of vision-language prediction is represented as the loss for next-token prediction:

\begin{linenomath}
\begin{equation}
    \mathcal{L}_{\text{traj}} =-\sum_{n=1}^{N} \sum_{i=1}^{C} y_{i,n} \log(\hat{y}_{i,n}),
\end{equation}
\end{linenomath}
where $N$ is the total number of tokens in the generated sequence, $C$ is the number of possible classes in the model's vocabulary, $y_{i,n}$ is a binary indicator indicating whether the $i$-th token is the correct one at the $n$-th position in the true sequence, $\hat{y}_{i,n}$ represents the predicted probability of the $i$-th token at the $n$-th position in the predicted sequence.

\paragraph{Contrastive Alignment Loss.} 
The multimodel feature alignment is controlled by the image-text contrastive loss computed over similarity scores:

\begin{linenomath}
\begin{equation}
    \mathcal{L}_{\text{align}} = -\frac{1}{K}\sum_{i=1}^K \log\frac{\exp(S_{ii})}{\sum_{j=1}^K \exp(S_{ij})}.
\end{equation}
\end{linenomath}

\paragraph{Knowledge Distillation Loss.} 
The KL divergence loss measures the discrepancy between the student's predictions and the teacher's softened distribution. Expanding the KL divergence term, we have:

\begin{linenomath}
\begin{equation}
    \mathcal{L}_{\text{KD}} = \mathcal{T}^2 \cdot \sum_{t=1}^T p_T^{(t)} \left(\log p_T^{(t)} - \log p_S^{(t)}\right),
\end{equation}
\end{linenomath}
where \(p_T^{(t)}\) and \(p_S^{(t)}\) denote the probabilities of the teacher and the student at the point of the trajectory \(t\), respectively. 

\paragraph{Aggregated Objective.} 
The final training loss combines these components with the weighting factors \(\lambda_1\), \(\lambda_2\):

\begin{linenomath}
\begin{equation}
    \mathcal{L}_{\text{total}} = \mathcal{L}_{\text{traj}} + \lambda_1 \mathcal{L}_{\text{align}} + \lambda_2 \mathcal{L}_{\text{KD}}.
\end{equation}
\end{linenomath}

The full minibatch update routine and the corresponding on-board inference
workflow are provided in Appendix~\ref{app:algorithms} as
Algorithm~\ref{alg:train} and Algorithm~\ref{alg:inference}, respectively. The gradients of the alignment and KD losses, together with the full
layer-wise complexity analysis, are provided in Appendix~C.1-C.4.

\section{Experiments}

\begin{table}[ht]
    \centering
    \caption{Comparison of L2 error, collision rate, and transmission cost across different methods. Lower L2 error and collision rate indicate better planning accuracy and safety, while transmission cost reflects the required bandwidth in BPS.}
    \label{tab:results}
    \begin{tabular}{l|cccc|cccc|c}
    \toprule
    \multirow{2}{*}{Method} 
        & \multicolumn{4}{c|}{L2 Error (m) $\downarrow$}
        & \multicolumn{4}{c|}{Collision Rate (\%) $\downarrow$}
        & \multirow{2}{*}{Transmission Cost (BPS) $\downarrow$} \\
    & 2.5s & 3.5s & 4.5s & Avg & 2.5s & 3.5s & 4.5s & Avg & \\
    \midrule
    UniV2X - No Fusion & 2.58 & 3.37 & 4.36 & 3.44 & 0.15 & 1.04 & 1.48 & 1.08 & 0 \\
    UniV2X - Vanilla & 2.33 & 3.69 & 5.12 & 3.71 & 0.59 & 2.07 & 3.70 & 2.07 & $8.19 \times 10^7$ \\
    UniV2X - BEV Feature Fusion & 2.31 & 3.29 & 4.31 & 3.30 & 0.00 & 1.04 & 1.48 & 0.93 & $8.19 \times 10^7$ \\
    UniV2X (\cite{yu2024end}) & 2.59 & 3.35 & 4.49 & 3.48 & \textbf{0.00} & 0.44 & 0.59 & 0.34 & $\textbf{8.09} \mathbf{\times} \textbf{10}^\textbf{5}$ \\
    CooperNaut (\cite{hu2023collaboration}) & 3.84 & 5.33 & 6.87 & 5.35 & 0.44 & 1.33 & 1.93 & 0.54 & $8.19 \times 10^5$ \\
    \textbf{V2X-VLM (Ours)} & \textbf{1.09} & \textbf{1.12} & \textbf{1.42} & \textbf{1.21} & 0.02 & \textbf{0.03} & \textbf{0.03} & \textbf{0.03} & $1.24 \times 10^7$ \\
    \bottomrule
    \end{tabular}
\end{table}

\subsection{Dataset}
The proposed V2X-VLM framework is evaluated on the DAIR-V2X dataset (\cite{yu2022dair}), an extensive and well-annotated resource designed for research on cooperative autonomous driving V2X. It includes 22,325 frames of data from vehicle-mounted sensors and 10,084 frames from infrastructure sensors, capturing RGB images and LiDAR data at up to 25 Hz. This comprehensive dataset is crucial for tasks such as trajectory prediction and multi-sensor data fusion, which facilitates the development of V2X systems that improve traffic safety, navigation accuracy, and cooperative driving strategies.

\subsection{Implementation Details}
\label{sec:hyperparamenters}
We implement the proposed V2X-VLM framework using PyTorch and train it on a single NVIDIA RTX 4090 GPU. We use Florence-2 (\cite{xiao2024florence}) as the VLM backbone. Florence-2 is one of state-of-the-art VLMs that delivers high-quality multimodal representations and fine-grained visual understanding. Specifically, the Florence-2-large model trained serves as the teacher model, while the Florence-2-base model serves as the student. During fine-tuning, the vision encoder parameters in the student model are kept frozen to ensure efficient learning. Training converges in 10 epochs using the AdamW optimizer with a batch size of 4, a learning rate of $1\times10^{-6}$ and a linear learning rate scheduler. The loss of contrastive alignment and the loss of knowledge distillation are weighted by hyperparameter $\lambda_1=0.1$ and $\lambda_2=0.5$, respectively. The distillation employs KL divergence with a temperature scaling factor of $\mathcal{T}=2.0$. The derivation that justifies this constant $\mathcal{T}$ choice is shown in Appendix C.3-C.4. The planning results of V2X-VLM are assessed using the metrics of L2 error, collision rate, and the transmission cost.

\subsection{Results Evaluation}

\subsubsection{L2 Error and Collision Rate}

Table~\ref{tab:results} compares the performance of the cooperative autonomous driving methods of baseline in terms of L2 error, collision rate, and transmission cost. UniV2X (\cite{yu2024end}) develops a state-of-the-art E2E pipeline that fuses data from both vehicles and infrastructure to improve perception, online mapping and planning. However, while UniV2X pioneers an E2E vehicle-infrastructure cooperative autonomous driving (VICAD) framework, the L2 errors of trajectory planning and collision rates remain higher than those of our proposed approach. CooperNaut (\cite{hu2023collaboration}) uses a decentralized fusion strategy for V2V communication to improve perception, but it still fails in trajectory planning and safety compared to our method. In contrast, V2X-VLM achieves the lowest L2 error across all time horizons and maintains an average low collision rate of 0.03\%. These improvements can be attributed to the advanced multimodal fusion of vehicle and infrastructure imagery with textual scene descriptions, complemented by contrastive learning and knowledge distillation to further refine feature alignment.

\subsubsection{Transmission Cost}

\begin{table}[htbp]
    \centering
    \caption{Performance comparison for different downsampling scaling factors in V2X communication.}
    \renewcommand{\arraystretch}{1.3}
    \setlength{\tabcolsep}{4pt} 
    \resizebox{\textwidth}{!}{ 
    \begin{tabular}{c|c|m{5.8cm}<{\centering}|c|c c c c|c|c}
        \toprule
        \multirow{2}{*}{Scaling Factor} & 
        \multicolumn{1}{c|}{\multirow{2}{*}{\makecell{Resolution after \\ DownSampling}}} & 
        \multirow{2}{*}{\centering Information Quality} & 
        \multirow{2}{*}{\makecell{Transmission \\ Cost (BPS) $\downarrow$}} & 
        \multicolumn{4}{c|}{L2 Error (m) $\downarrow$} & 
        \multirow{2}{*}{\makecell{Total \\ Latency (ms)}} & 
        \multirow{2}{*}{FPS} \\
        & & & & 2.5s & 3.5s & 4.5s & Avg. & & \\
        \midrule
        UniV2X & - & - & $8.09 \times 10^5$ & 2.59 & 3.35 & 4.49 & 3.48 & - & - \\
        \midrule
        1 (No Change) & $1080 \times 1920$ & Full resolution; all details preserved; highest fidelity. & $1.24 \times 10^7$ & \textbf{1.09} & \textbf{1.12} & \textbf{1.42} & \textbf{1.21} & 353.36 & 11.32 \\
        \midrule
        0.5 & $540 \times 960$ & Moderate downsampling; acceptable detail with minor loss in fine features; moderate fidelity & $3.11 \times 10^6$ & 1.23 & 1.27 & 1.45 & 1.32 & 352.90 & 11.33 \\
        \midrule
        0.2 & $316 \times 384$ & High downsampling; significant reduction in detail; degraded fidelity & $4.98 \times 10^5$ & 1.34 & 1.38 & 1.59 & 1.44 & 264.79 & 15.11 \\
        \midrule
        0.1 & $108 \times 192$ & High downsampling; severe loss of details; low fidelity & $\textbf{1.24} \mathbf{\times} \textbf{10}^\textbf{5}$ & 1.42 & 1.47 & 1.71 & 1.53 & \textbf{263.97} & \textbf{15.15} \\
        \bottomrule
    \end{tabular}
    }
    
    \label{tab:transmission_tradeoff}
\end{table}

In V2X-VLM, cooperative perception is achieved by integrating both vehicle-side and infrastructure-side images into the VLM backbone. Since infrastructure-side images are not locally available on the ego vehicle, they must be transmitted over a communication network before being processed. This introduces a trade-off between transmission cost and planning accuracy. The method of calculating transmission cost is provided in Appendix A. When transmitting full-resolution images at \(1080 \times 1920\), the required bandwidth reaches \(1.24 \times 10^7\) BPS, significantly higher than the \(8.09 \times 10^5\) BPS used in UniV2X, which relies on feature-level transmission rather than raw image sharing. This substantial transmission cost arises from the direct transmission of high-resolution images, as the bandwidth requirement scales with pixel density.

To mitigate the high communication overhead, a practical approach is to downsample the infrastructure-side images before transmission and upsample them upon reception. This reduces the amount of data sent over the network while still allowing the VLM to process the visual information. As shown in Table~\ref{tab:transmission_tradeoff}, a scaling factor of 0.5 decreases the transmission cost to \(3.11 \times 10^6\) BPS, while extreme downsampling with 0.1 scaling factor reduces it to just \(1.24 \times 10^5\) BPS, achieving a two-order-of-magnitude reduction. However, a lower resolution leads to inevitable degradation in visual detail, which impacts performance. The L2 error increases from 1.21 m at full resolution to 1.53 m at the lowest resolution, highlighting the trade-off between bandwidth efficiency and accuracy.

Beyond transmission cost and planning accuracy, there are additional trade-offs to consider. As image resolution decreases, computational efficiency improves due to lower input complexity. This effect is evident in the metrics of FPS and per-batch inference latency: the highest-resolution setup runs at 11.32 FPS with a latency of 353.36 ms, whereas the lowest-resolution configuration achieves 15.15 FPS with a reduced latency of 263.97 ms. These results suggest that aggressive compression benefits real-time inference while introducing a slight accuracy penalty. Notably, despite using an extremely low-resolution infrastructure image, V2X-VLM still consistently outperforms UniV2X in trajectory planning across all time horizons. This demonstrates that the proposed multimodal feature alignment and VLM-based reasoning can effectively compensate for degraded visual information, ensuring robust trajectory prediction even under constrained bandwidth conditions.


\subsection{Robustness and Efficiency Analysis}


\begin{table*}[ht]
    \centering
    \caption{Robustness evaluation of V2X-VLM under perturbations.}
    \label{tab:robustness}
    \begin{tabular}{l|ccc|ccc}
    \toprule
    \multirow{2}{*}{Condition} 
        & \multicolumn{3}{c|}{L2 Error (m) $\downarrow$} 
        & \multicolumn{3}{c}{Collision Rate (\%) $\downarrow$} \\
         & 2.5s & 3.5s & 4.5s & 2.5s & 3.5s & 4.5s \\
    \midrule
    Image Noise (std = 5)       & 1.31 & 1.34 & 1.50 & 0.03 & 0.03 & 0.04 \\
    Image Noise (std = 10)      & 1.17 & 1.21 & 1.49 & 0.03 & 0.03 & 0.04 \\
    Text Perturbation (p = 0.1)   & 1.33 & 1.37 & 1.46 & 0.02 & 0.03 & 0.04 \\
    \makecell[l]{Combined (Image Noise 10, Text p = 0.1)} 
                                & 1.34 & 1.36 & 1.76 & 0.02 & 0.03 & 0.03 \\
    \midrule
    No Perturbation             & \textbf{1.09} & \textbf{1.12} & \textbf{1.42} & \textbf{0.02} & \textbf{0.03} & \textbf{0.03} \\
    \bottomrule
    \end{tabular}%
\end{table*}

To assess the robustness of V2X-VLM, we introduce controlled perturbations to both the visual and textual inputs and evaluate their effects on trajectory accuracy and safety. Table~\ref{tab:robustness} shows that the model maintains strong performance even under perturbed conditions. Adding Gaussian noise to infrastructure-side images (\textit{Image Noise}) slightly increases the L2 error but does not significantly degrade planning accuracy, as seen in the cases of standard deviation 5 and 10. Text perturbation (\textit{Text Perturbation}) simulates potential errors in language descriptions by randomly modifying portions of textual inputs. The impact on L2 error remains minor, highlighting the robustness of the model in handling imperfect textual descriptions. When both image and text perturbations are applied simultaneously (\textit{Combined Perturbation}), the average L2 error increases to 1.49, still outperforming existing baselines shown in Table~\ref{tab:results}. The collision rate remains nearly constant across all perturbation settings, further demonstrating V2X-VLM’s stability and resilience to noisy inputs.


\begin{table}[ht]
    \centering
    \footnotesize
    \caption{Latency breakdown and FPS analysis for inference efficiency. Total latency represents the time taken for a batch of inputs. FPS indicates the number of frames processed per second.}
    \label{tab:latency}
    \begin{tabular}{llcc}
    \toprule
    Process & Description & Latency (ms) & Proportion (\%) \\
    \midrule
    Preprocessing  & \makecell[l]{Tokenization and image processing}          & 269.01 & 76.1 \\
    Inference      & \makecell[l]{Forward pass through the model}               & 72.72  & 20.6 \\
    Postprocessing & \makecell[l]{Decoding the model outputs}                    & 9.02   & 2.6  \\
    \makecell[l]{Residual\\ Overhead} & \makecell[l]{Minor operations such as data loading, \\synchronization, and loop overhead} & 2.61 & 0.7 \\
    \midrule
    Total          & -                                             & 353.36 & 100.0 \\
    \midrule
    FPS            & -                                             & 11.32  & -    \\
    \bottomrule
    \end{tabular}
\end{table}


Beyond robustness, we analyze the inference efficiency of V2X-VLM by breaking down the total latency per batch in Table~\ref{tab:latency}. The overall latency for a batch of samples is 353.36 ms, corresponding to a real-time processing rate of 11.32 FPS. The majority of the latency stems from preprocessing (76.1\%), which includes tokenization and image feature extraction. This step is necessary for the multimodal input fusion but could be further optimized. The model’s forward pass (\textit{Inference}) accounts for 20.6\% of the total latency, reflecting the computational cost of large-scale vision language processing. \textit{Postprocessing}, which involves decoding the model output into trajectories, is relatively lightweight (2.6\%), and the residual overhead from data loading and synchronization is negligible (0.7\%). Despite computational complexity, V2X-VLM achieves real-time inference capabilities, demonstrating its feasibility for deployment in practical cooperative autonomous driving systems.

\begin{table*}[ht]
    \centering
    \caption{Ablation study result. Removing each component of V2X-VLM degrades planning accuracy, demonstrating their importance.}
    \label{tab:ablation}
    \begin{tabular}{l|cccc|cccc|c}
        \toprule
        \multirow{2}{*}{Method} 
        & \multicolumn{4}{c|}{L2 Error (m) $\downarrow$}
        & \multicolumn{4}{c|}{Collision Rate (\%) $\downarrow$}
        & \multirow{2}{*}{Transmission Cost (BPS) $\downarrow$} \\
        & 2.5s & 3.5s & 4.5s & Avg. & 2.5s & 3.5s & 4.5s & Avg. & \\
        \midrule
        No Fusion & 1.45 & 1.50 & 1.53 & 1.49 & 0.03 & 0.03 & 0.04 & 0.03 & \textbf{0} \\
        w/o Distillation & 1.33 & 1.33 & 1.59 & 1.42 & 0.03 & 0.03 & 0.03 & 0.03 & $1.24 \times 10^7$ \\
        w/o Scene Prompting & 1.34 & 1.37 & 1.58 & 1.43 & 0.03 & 0.03 & 0.03 & 0.03 & $1.24 \times 10^7$ \\
        w/o Feature Alignment & 1.40 & 1.44 & 1.69 & 1.51 & 0.03 & 0.03 & 0.04 & 0.03 & $1.24 \times 10^7$ \\
        \textbf{V2X-VLM (Ours)} & \textbf{1.09} & \textbf{1.12} & \textbf{1.42} & \textbf{1.21} & \textbf{0.02} & \textbf{0.03} & \textbf{0.03} & \textbf{0.03} & $1.24 \times 10^7$ \\
        \bottomrule
    \end{tabular}
\end{table*}

\begin{figure*}[ht]
    \centering
    \includegraphics[width=\textwidth]{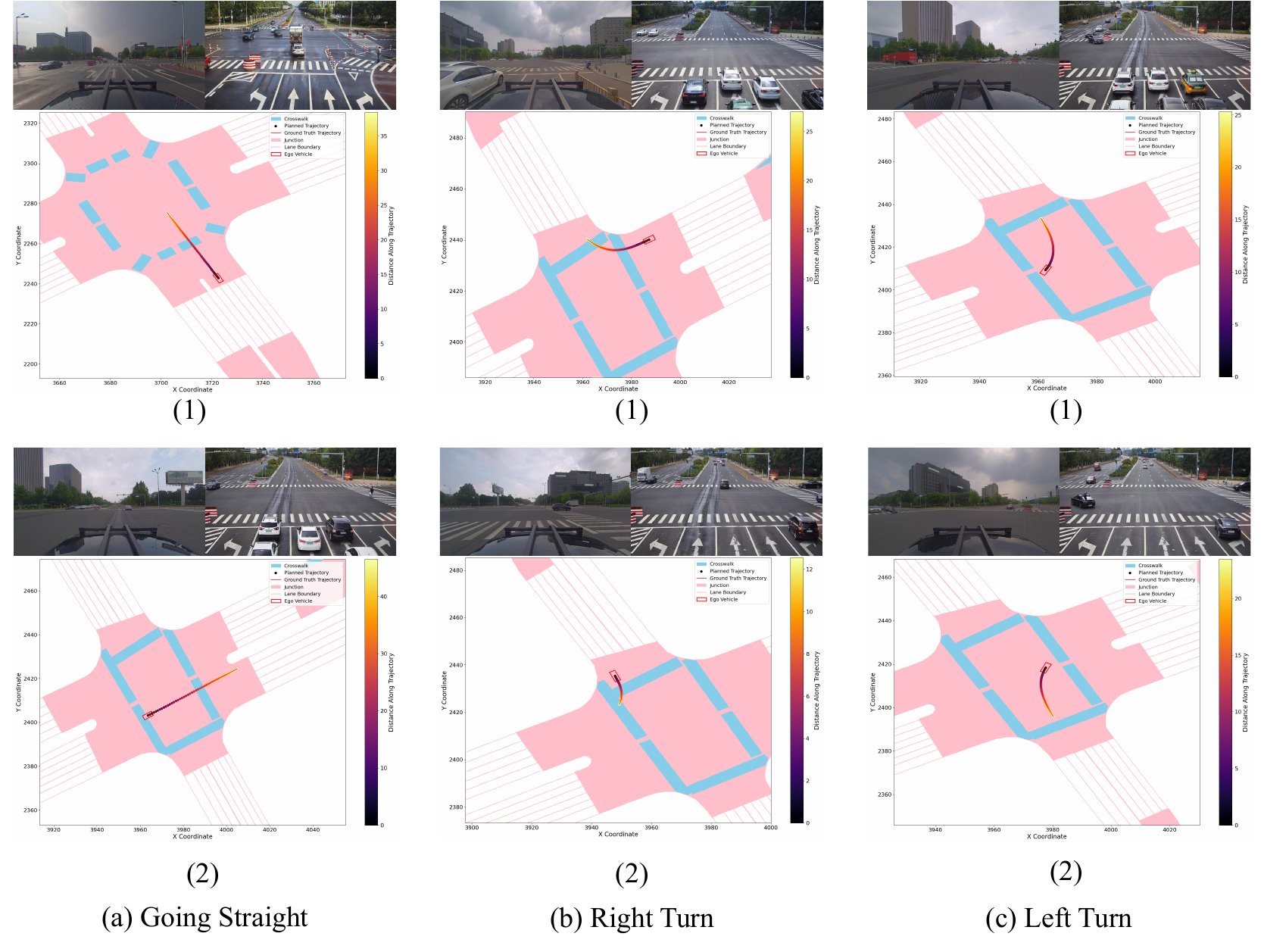}
    \caption{Visualization of V2X-VLM trajectory planning on three common driving scenarios. Continuous frames are visualized at a frequency of 1 Hz.}
    \label{fig:normal_scenarios}
\end{figure*}

\subsection{Ablation Study}

We conduct an ablation study to evaluate the contributions of key components in V2X-VLM. Results are presented in Table~\ref{tab:ablation}. Each ablation setting is described as follows:

\begin{itemize}
\item \textbf{No Fusion:} Only ego-vehicle images are used, omitting infrastructure-side input.
\item \textbf{w/o Knowledge Distillation:} The student model is trained without knowledge distillation.
\item \textbf{w/o Scene Prompting:} The semantic textual scene descriptions are removed from the input.
\item \textbf{w/o Feature Alignment:} The contrastive learning-based feature alignment between image and text is disabled.
\end{itemize}

As shown in Table~\ref{tab:ablation}, removing infrastructure input results in the highest L2 error, demonstrating the necessity of multiperspective fusion. This finding reinforces the fundamental advantage of cooperative autonomous driving over a single-vehicle-based solution. The significant performance drop in the single vehicle setting underscores the limitations of independent perception and planning, further supporting the case for cooperative autonomous driving. Among the model components, the alignment of the features contributes the most to accuracy, followed by the prompting of the scene and the distillation of knowledge, highlighting their role in multimodal understanding and trajectory planning. The results validate the effectiveness of each component and demonstrate their combined impact in achieving state-of-the-art performance.

\subsection{Visualization}

To further illustrate the effectiveness of V2X-VLM, we provide qualitative results showcasing the performance of V2X-VLM in both normal scenarios and some corner cases.

Figure~\ref{fig:normal_scenarios} showcases the trajectories planned by V2X-VLM in three common driving maneuvers: left turn, going straight, and right turn. It illustrates the consistent ability of V2X-VLM to produce high-quality trajectory output. Further visualization and discussion regarding more challenging corner cases, such as rainy conditions with vehicle camera lens blur, as well as driving through complex intersections, are presented in Appendix D.




\section{Conclusion}
This paper introduces V2X-VLM, an end-to-end cooperative autonomous driving framework that integrates multimodal scene understanding using VLMs for enhanced trajectory planning. By fusing multiperspective vehicle and infrastructure images with semantic text and leveraging contrastive feature alignment alongside knowledge distillation, the model achieves state-of-the-art planning accuracy over baseline methods. The ablation study validates the necessity of each component. Robustness evaluations further validate the model’s resilience against input perturbations, while efficiency analysis highlights its feasibility for real-time deployment.


Future work will focus on two key areas of improvement. First, we aim to enhance the model's generalization by addressing more long-tail scenarios. This will involve generating various long-tail scenarios for model training and evaluation. Second, efforts will be made to reduce transmission costs by exploring a vehicle-road-cloud distributed training and deployment paradigm dedicated to optimizing the balance between data processing and communication for scalable and cost-effective driving applications.

\appendix

\section{Communication Cost Calculation}

In V2X cooperative perception, infrastructure-side images must be transmitted over a network before processing, which introduces a significant communication overhead. The transmission cost, measured in bytes per second (BPS), depends on image resolution, color channels, transmission frequency, and potential downsampling.

For an image of width \( W \), height \( H \), and \( C \) color channels, transmitted at frequency \( f \), the required bandwidth is:
\begin{equation}
    \text{BPS} = s^2 W H C f,
\end{equation}
where \( s \) is the downsampling factor (\( 0 < s \leq 1 \)), reducing the image resolution before transmission.

Following this, for a full-resolution infrastructure-side image of \( 1080 \times 1920 \) pixels with three color channels, transmitted at 2 Hz without downsampling (\( s=1 \)), the required bandwidth is calculated as $1.24 \times 10^7 \text{ BPS}$.

\section{Core Training and Inference Implementation}
\label{app:algorithms}

This appendix consolidates the two key procedural components of our
\mbox{V2X-VLM} framework: the \emph{online cooperative inference pipeline},
which runs on the ego vehicle in real time, and the \emph{student–teacher
training routine}, used offline to optimise the lightweight student model.
The pseudocode is intentionally concise to highlight
where the critical processes, such as V2X communication, multimodal reasoning, and trajectory planning, occur during inference, and how alignment and knowledge–distillation (KD)
losses are combined during training.

\begin{algorithm}[H]
\caption{Real-Time Cooperative Inference Flow}
\label{alg:inference}
\begin{algorithmic}[1]
\Require Ego image $I_v$, roadside image $I_i$, pre-trained foundation model $\Phi$
\Ensure Smoothed future trajectory $\tau$
\State \textbf{Spawn Task-A (vehicle)}: Generate scene text $E=\Phi_{\text{describe}}(I_v)$  \Comment{e.g.\ GPT-4o, florence-2-large}
\State \textbf{Spawn Task-B (roadside)}: Compress \textbf{(Optional)}, and send $I_i$ as payload $P_i$
\State \textbf{Synchronize tasks (vehicle)}: Receive roadside image $I_i$ (decode if compressed)  \Comment{ensure $E$ is ready}
\State \textbf{Run V2X-VLM}: Compute multimodal feature $f=\text{V2X\!-\!VLM}(I_{\text{mv}}=[I_v,I_i],E)$
\State \textbf{Decode trajectory tokens}: Obtain $\hat{\tau}=\mathrm{Decoder}(f)$
\State \textbf{Apply refinement}: Produce $\tau=\mathrm{Refine}(\hat{\tau})$
\State \Return $\tau$
\end{algorithmic}
\end{algorithm}

\begin{algorithm}[H]
\caption{Offline Training with Alignment \& Knowledge Distillation}
\label{alg:train}
\begin{algorithmic}[1]
\Require Pre-trained teacher foundation model $F_T$, student V2X-VLM model, training batch $\mathcal{B}$
\Ensure Updated student parameters $\theta_S$
\ForAll{$(I_v,I_i,\tau^*) \in \mathcal{B}$}
    \State \textbf{Generate preliminary scene description text}: $E_0=F_T(I_v)$  \Comment{e.g.\ florence-2-large}
    \State \textbf{Refine text manually}: Obtain final $E$  \Comment{crowd-worker QA}
    \State \textbf{Forward teacher model}: Predict $\tau_T = F_T(I_v,I_i,E)$
    \State \textbf{Forward student model}: Predict $\tau_S , z , h = \text{V2X-VLM}(I_v,I_i,E)$
    \State \textbf{Calculate alignment loss}: Compute $\mathcal{L}_{\text{align}}$ via InfoNCE \Comment{Eq.\,10}
    \State \textbf{Calculate KD loss}: Compute $\mathcal{L}_{\text{KD}}$ with temperature $\mathcal{T}$ \Comment{Eq.\,8}
    \State \textbf{Calculate trajectory loss}: Compute $\mathcal{L}_{\text{traj}}$ for next-token prediction \Comment{Eq.\,9}
\EndFor
\State \textbf{Aggregate losses}: Obtain $\mathcal{L}_{\text{total}}
       = \mathcal{L}_{\text{traj}}
       + \lambda_1 \mathcal{L}_{\text{align}}
       + \lambda_2 \mathcal{L}_{\text{KD}}$  \Comment{Eq.\,12}
\State \textbf{Update parameters}: Apply AdamW to $\theta_S$
\end{algorithmic}
\end{algorithm}

Algorithms~\ref{alg:inference} and~\ref{alg:train} are referenced in
Sections \ref{sec:method}–\ref{sec:training} of the main text for clarity.
They require no additional hyperparameters beyond those described in Section~\ref{sec:hyperparamenters}.

\section{Theoretical Analysis}
\label{app:analysis}

This appendix expands the concise derivations given in the main paper.
The first part details the floating-point operations (FLOPs) for a multimodal transformer layer.
The second part derives the gradients of the image–text alignment loss and the knowledge-distillation (KD) loss.
The third part analyses how the temperature \(\mathcal{T}\) influences optimization.

\subsection{Exact FLOP Count per Layer}
\label{app:complexity-detailed}

The FLOPs for a single multimodal transformer layer are derived from the standard
\(Q\,K^{\!\top}\!V\) formulation.
Let the hidden size be \(d\),
the number of heads be \(h\) such that the head dimension satisfies \(d_h = d/h\),
the sequence length for the vision branch be \(N_v\),
and the sequence length for the text branch be \(N_t\).

The vision branch self-attention requires
\begin{equation}
\text{FLOPs}_{\text{vis}}
= 2N_v d^2 + N_v^{2}d,
\label{eq:vis-flops}
\end{equation}
where the first term is for the linear projections and the second term covers
\(QK^{\!\top}\) and the subsequent \(\text{softmax}(QK^{\!\top})V\).

Likewise, the text branch self-attention requires
\begin{equation}
\text{FLOPs}_{\text{text}}
= 2N_t d^2 + N_t^{2}d.
\label{eq:text-flops}
\end{equation}

The cross-modal attention, taking the vision tokens as query and the text tokens as key–value, requires
\begin{equation}
\text{FLOPs}_{\text{cross}}
= 2N_v d^2 + N_v N_t d.
\label{eq:cross-flops}
\end{equation}

Since $d$ is fixed after architecture selection, the projection terms
$2N_v d^{2}$ and $2N_t d^{2}$ grow only linearly in the token numbers
and are therefore asymptotically dominated by the attention terms.
Discarding these lower-order components yields
\begin{equation}
\mathcal{O}(N_v d)+\mathcal{O}(N_t d)+\boxed{\mathcal{O}(N_v N_t)},
\label{eq:complexity}
\end{equation}

which shows that cross-modal attention is
the only \(\Theta(N^2)\) bottleneck.

Although the main experiments in Section 5 keep the full-rank cross-modal projections, for deployment on resource-constrained hardware, each \(d\times d\) projection can be replaced by a low-rank product
\(AB^{\!\top}\) with \(A,B\in\mathbb{R}^{d\times r}\).
This change reduces the two projection terms in
\(\text{FLOPs}_{\text{cross}}\) from \(2d^{2}\) to \(2dr\)
and leaves the rest of the layer untouched.
Detailed derivations appear later in this subsection,
and an example with \(r=4\) yields a four-to-one reduction in these
projections. The adapter is therefore a compatible but optional
engineering choice rather than part of the core model.

\subsection{Gradient of the Alignment Loss}
\label{app:align-grad-detailed}

The InfoNCE alignment loss for a batch of size \(K\) is defined as
\[
\mathcal{L}_{\text{align}}
= - \frac{1}{K}\sum_{i=1}^{K}\log
   \frac{\exp(S_{ii})}{\sum_{j=1}^{B}\exp(S_{ij})},
\quad
S_{ij}=\langle z_i, h_j\rangle / \kappa.
\]
Let
\(
\sigma_{ij} = \exp(S_{ij}) / \sum_{k}\exp(S_{ik})
\), and then

\begin{empheq}[box=\fbox]{align}
\frac{\partial\mathcal{L}_{\text{align}}}{\partial S_{ii}}
&= -\frac{1}{K}(1 - \sigma_{ii}) \label{eq:align_pos_grad} \\
\frac{\partial\mathcal{L}_{\text{align}}}{\partial S_{ij}}
&= \frac{1}{K} \sigma_{ij} \quad j \neq i. \label{eq:align_neg_grad}
\end{empheq}

Equations~\eqref{eq:align_pos_grad} and~\eqref{eq:align_neg_grad}
show that the gradient becomes small once the model assigns high confidence to the true pair,
thereby concentrating learning on hard positives.

\subsection{Gradient of the Distillation Loss}
\label{app:kd-grad-detailed}

The temperature-scaled KD loss is
\[
\mathcal{L}_{\text{KD}}
= \frac{\mathcal{T}^{2}}{N}
  \sum_{k} p_T^{(k)} \log
  \frac{p_T^{(k)}}{p_S^{(k)}},
\quad
p_S = \text{softmax}\!\bigl(\tau_S/\mathcal{T}\bigr),
\;
p_T = \text{softmax}\!\bigl(\tau_T/\mathcal{T}\bigr).
\]
Since
\(\partial\log p_S^{(k)}/\partial\tau_S^{(\ell)}
  = (\delta_{k\ell} - p_S^{(\ell)})/\mathcal{T}\),
we have
\begin{equation}
\boxed{\;
\frac{\partial\mathcal{L}_{\text{KD}}}{\partial \tau_S^{(\ell)}}
= \frac{\mathcal{T}^{2}}{N}\bigl(p_S^{(\ell)} - p_T^{(\ell)}\bigr)
\;,}
\label{eq:kd_grad}
\end{equation}
which shows that the gradient scale grows with \(\mathcal{T}^{2}\).

\subsection{Impact of Temperature on Optimization}
\label{app:schedule-temp}

\paragraph{Temperature Scheduling}  
Equation \eqref{eq:kd_grad} shows that a sharp teacher distribution
reduces gradient magnitude at the start of training.
A constant temperature of \(\mathcal{T}=2\) amplifies the signal by a
factor of four throughout all epochs, which we found sufficient for
stable convergence and did not observe further benefit from annealing.

\paragraph{Combined Effect with Alignment Loss}  
The InfoNCE gradient concentrates on the hardest image–text pairs,
while the higher temperature in knowledge distillation restores the
signal that would otherwise vanish early.
Together these two mechanisms smooth the loss landscape
and shorten convergence time.

\paragraph{Summary of Findings}  
The boxed equations
\eqref{eq:complexity}, \eqref{eq:align_pos_grad}, \eqref{eq:align_neg_grad} and \eqref{eq:kd_grad}
justify the hyperparameter choice \(\mathcal{T}=2\),
which gives the best balance between computational efficiency
and prediction accuracy across all experiments.

\section{Corner Case Visualization}

\begin{figure}[ht]
    \centering
    \includegraphics[width=0.9\textwidth]{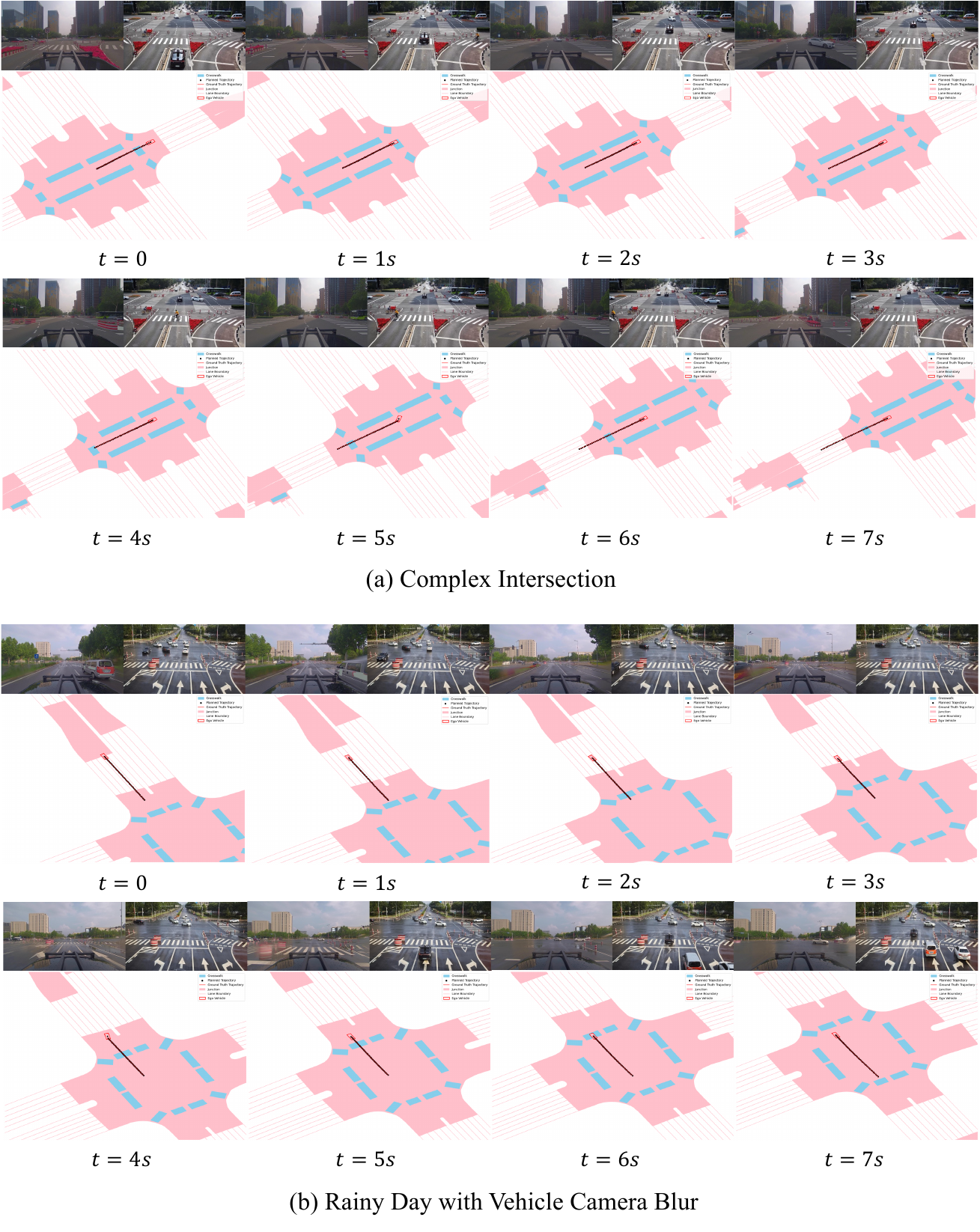}
    \caption{Visualization of V2X-VLM's trajectory planning for going-straight scenarios in challenging corner cases. Continuous frames are displayed at a frequency of 1 Hz.}
    \label{fig:fig5}
\end{figure}

\begin{figure}[ht]
    \centering
    \includegraphics[width=0.9\textwidth]{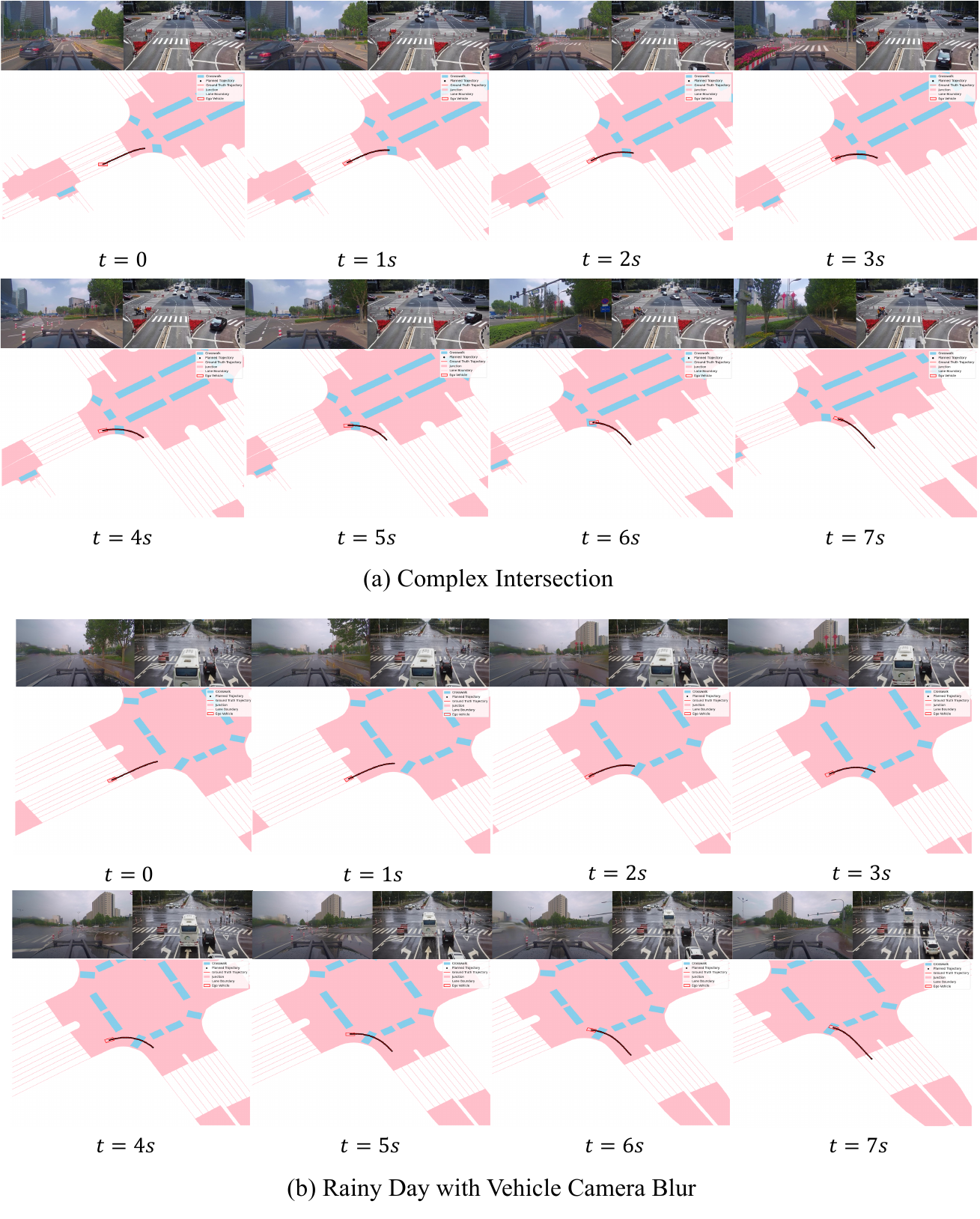}
    \caption{Visualization of V2X-VLM's trajectory planning for right-trun scenarios in challenging corner cases. Continuous frames are displayed at a frequency of 1 Hz.}
    \label{fig:fig4}
\end{figure}

\begin{figure}[ht]
    \centering
    \includegraphics[width=0.9\textwidth]{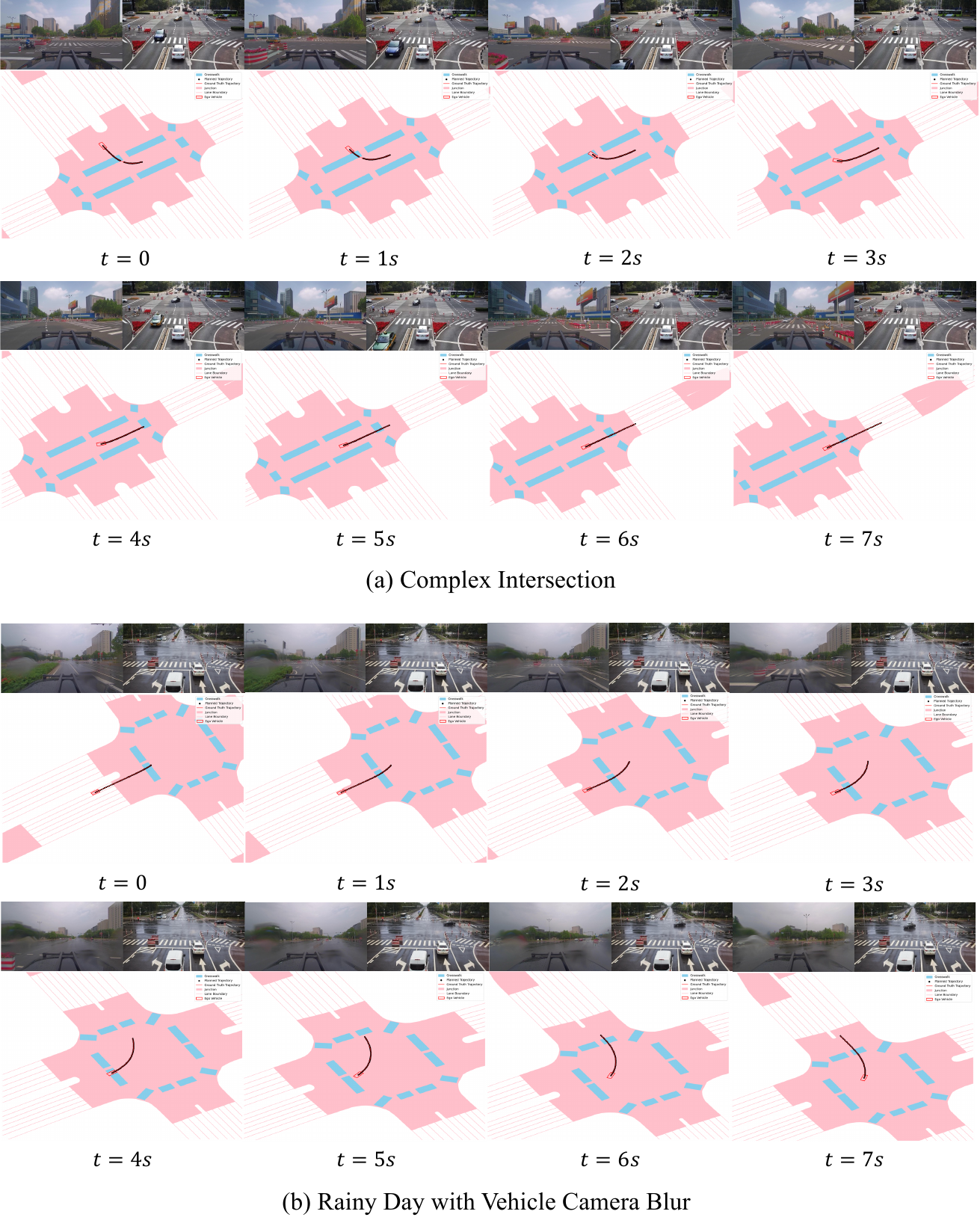}
    \caption{Visualization of V2X-VLM's trajectory planning for left-turn scenarios in challenging corner cases. Continuous frames are displayed at a frequency of 1 Hz.}
    \label{fig:fig6}
\end{figure}

\begin{figure}[ht]
    \centering
    \includegraphics[width=0.65\textwidth]{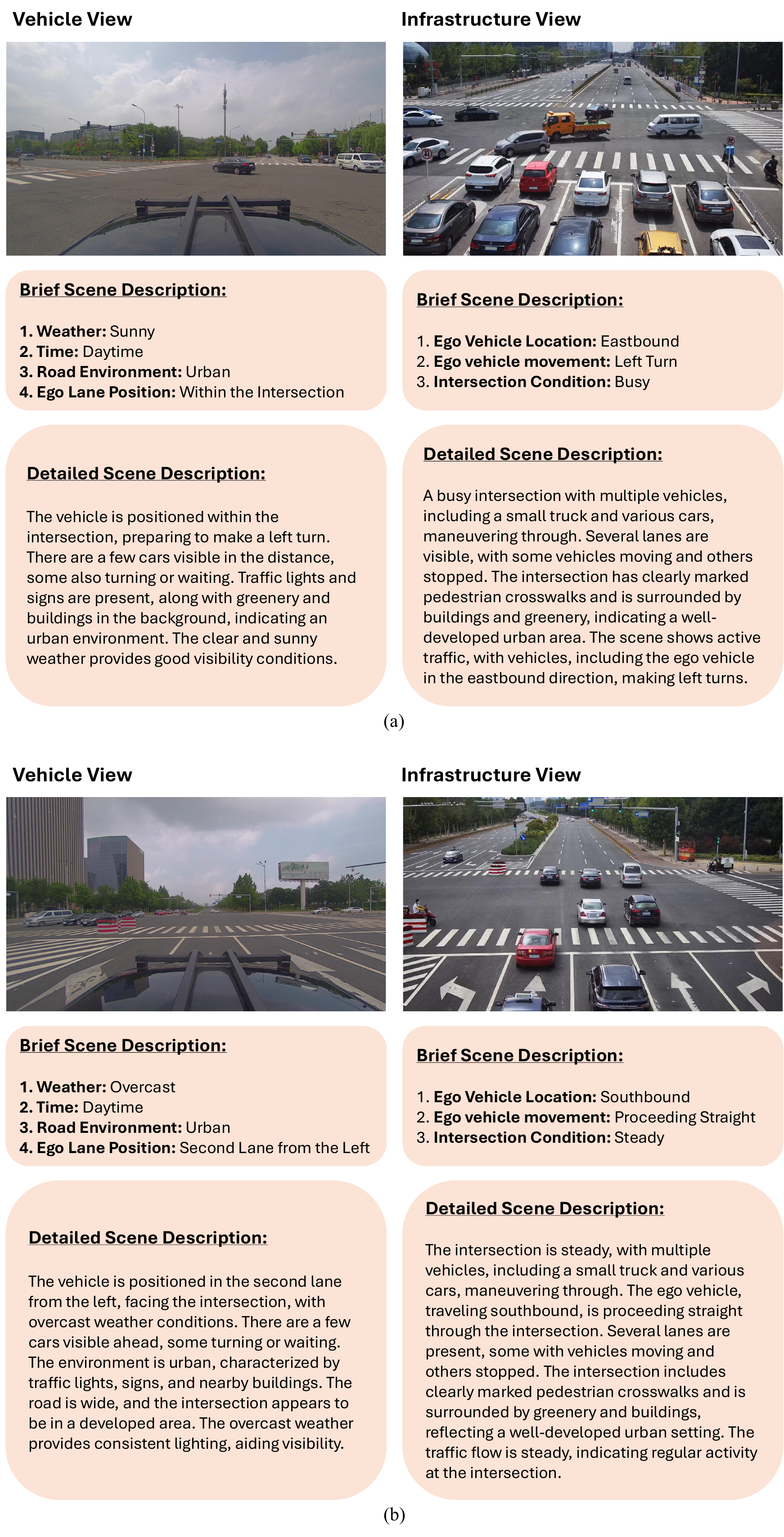} 
    \caption{Examples of VLM vehicle-side and infrastructure-side scene understanding.}
    \label{fig:text_prompt_examples}
\end{figure}

To further assess the robustness of V2X-VLM, we provide additional qualitative visualizations focusing on corner cases, including adverse weather conditions where vehicle camera lens is blur and complex intersection scenarios. Again, we examine three most common maneuvers, going straight, left turn, and right turn, under these challenging conditions. 

In adverse weather scenarios, blurred vehicle cameras reduce visibility, making it harder to extract reliable visual cues. This tests V2X-VLM’s ability to leverage infrastructure-side views and textual descriptions for robust trajectory planning. Meanwhile, complex intersections introduce ambiguous right-of-way situations and occlusions that require effective multiperspective and multimodal fusion to ensure accurate and safe navigation. The qualitative results, presented in Figure~\ref{fig:fig5}, Figgure~\ref{fig:fig4}, and Figure~\ref{fig:fig6}, demonstrate that V2X-VLM consistently generates stable and contextually appropriate trajectories despite these challenges, reinforcing its capability to handle real-world uncertainties in cooperative autonomous driving.


\section{Scene Understanding in VLM for Cooperative Autonomous Driving}

This section showcases the capability of VLMs in understanding driving scenes within the cooperative autonomous driving setup. The structured textual descriptions presented in Figure~\ref{fig:text_prompt_examples} are derived from VLM-based scene interpretation and are crafted as semantic prompts for training V2X-VLM.

The structured textual inputs include the following components:

\begin{itemize}
    \item \textbf{Brief Scene Description:} Summarizes the fundamental scene attributes, including weather conditions, time of day, type of road environment, and the current position of the ego-vehicle.
    \item \textbf{Detailed Scene Description:} Provides a more comprehensive semantic interpretation, describing surrounding vehicles, traffic density, infrastructure elements, and interaction dynamics within the scene.
\end{itemize}

These VLM-generated descriptions offer a high-level semantic representation of the driving environment, which is integrated into V2X-VLM’s input alongside vehicle and infrastructure camera images. By incorporating multimodal inputs, the model gains a deeper contextual understanding of driving scenarios, improving cooperative end-to-end trajectory planning. This highlights the effectiveness of VLMs in extracting structured knowledge from visual data to facilitate enhanced decision-making in cooperative autonomous driving.



\printcredits
\clearpage

\bibliographystyle{cas-model2-names}

\bibliography{cas-refs}





\end{document}